\pdfoutput=1
\documentclass[11pt]{article}

\usepackage{emnlp2021}

\usepackage{amsmath,amsfonts,bm}

\def\eqref#1{equation~\ref{#1}}
\def\1{\bm{1}}

\DeclareMathAlphabet{\mathsfit}{\encodingdefault}{\sfdefault}{m}{sl}
\SetMathAlphabet{\mathsfit}{bold}{\encodingdefault}{\sfdefault}{bx}{n}

\def\gD{{\mathcal{D}}}

\def\gK{{\mathcal{K}}}
\def\gL{{\mathcal{L}}}

\def\gN{{\mathcal{N}}}

\def\gV{{\mathcal{V}}}

\usepackage{times}
\usepackage{latexsym}

\usepackage[T1]{fontenc}
\usepackage[utf8]{inputenc}

\usepackage{microtype}

\usepackage{hyperref}       % hyperlinks

\usepackage{url}            % simple URL typesetting
\usepackage{booktabs}       % professional-quality tables
\usepackage{amsfonts}       % blackboard math symbols
\usepackage{nicefrac}       % compact

\usepackage{amssymb} 
\usepackage{enumitem} 

\usepackage{subcaption}
\usepackage{caption}
\usepackage{titlesec}
\usepackage{multirow}
\usepackage{xcolor}
\usepackage{graphicx}
\usepackage{tabularx}
\usepackage{arydshln}
\usepackage{wrapfig}
\usepackage{bbm}
\usepackage{makecell}
\usepackage{sidecap}

\usepackage{algorithm}
\usepackage{algorithmicx, algpseudocode}

\usepackage[normalem]{ulem}

\usepackage{times}
\usepackage{latexsym}

\newcommand{\knnlm}{$k$NN-LM }

\title{Efficient Nearest Neighbor Language Models}

\author{Junxian He$^{\dagger}$, Graham Neubig$^{\dagger}$, Taylor Berg-Kirkpatrick$^{\ddagger}$ \\
  $^{\dagger}$Language Technologies Institute, Carnegie Mellon University \\
  $^{\ddagger}$Department of Computer Science and Engineering, University of California San Diego\\
  \texttt{\{junxianh, gneubig\}@cs.cmu.edu, tberg@eng.ucsd.edu}\\}

\begin{document}
\maketitle
\begin{abstract}
% \gn{For the name of the paper, what about ``Efficient Nearest Neighbor Language Models'' or ``Improving the Efficiency of Nearest Neighbor Language Models''?}
Non-parametric neural language models (NLMs) learn predictive distributions of text utilizing an external datastore, which allows them to learn through explicitly memorizing the training datapoints. While effective, these models often require retrieval from a large datastore at test time, significantly increasing the inference overhead and thus limiting the deployment of non-parametric NLMs in practical applications. In this paper, we take the recently proposed $k$-nearest neighbors language model~\citep{khandelwal2019generalization} as an example, exploring methods to improve its efficiency along various dimensions. Experiments on the standard WikiText-103 benchmark and domain-adaptation datasets show that our methods are able to achieve up to a 6x speed-up in inference speed while retaining comparable performance. The empirical analysis we present may provide guidelines for future research seeking to develop or deploy more efficient non-parametric NLMs.\footnote{Code is available at \href{https://github.com/jxhe/efficient-knnlm}{https://github.com/jxhe/efficient-knnlm}.} 
\end{abstract}

%!TEX root=./emnlp2021.tex
\section{Introduction}

Language models (LMs) are one of the most fundamental technologies in NLP, with applications spanning text generation~\citep{bahdanau2014neural,rush2015neural}, representation learning~\citep{peters2018deep,devlin2019bert,yang2019xlnet}, and few-shot learning~\citep{radford2019language,brown2020language}. 
\begin{figure}[!t]
\centering
    \includegraphics[width=1.0\columnwidth]{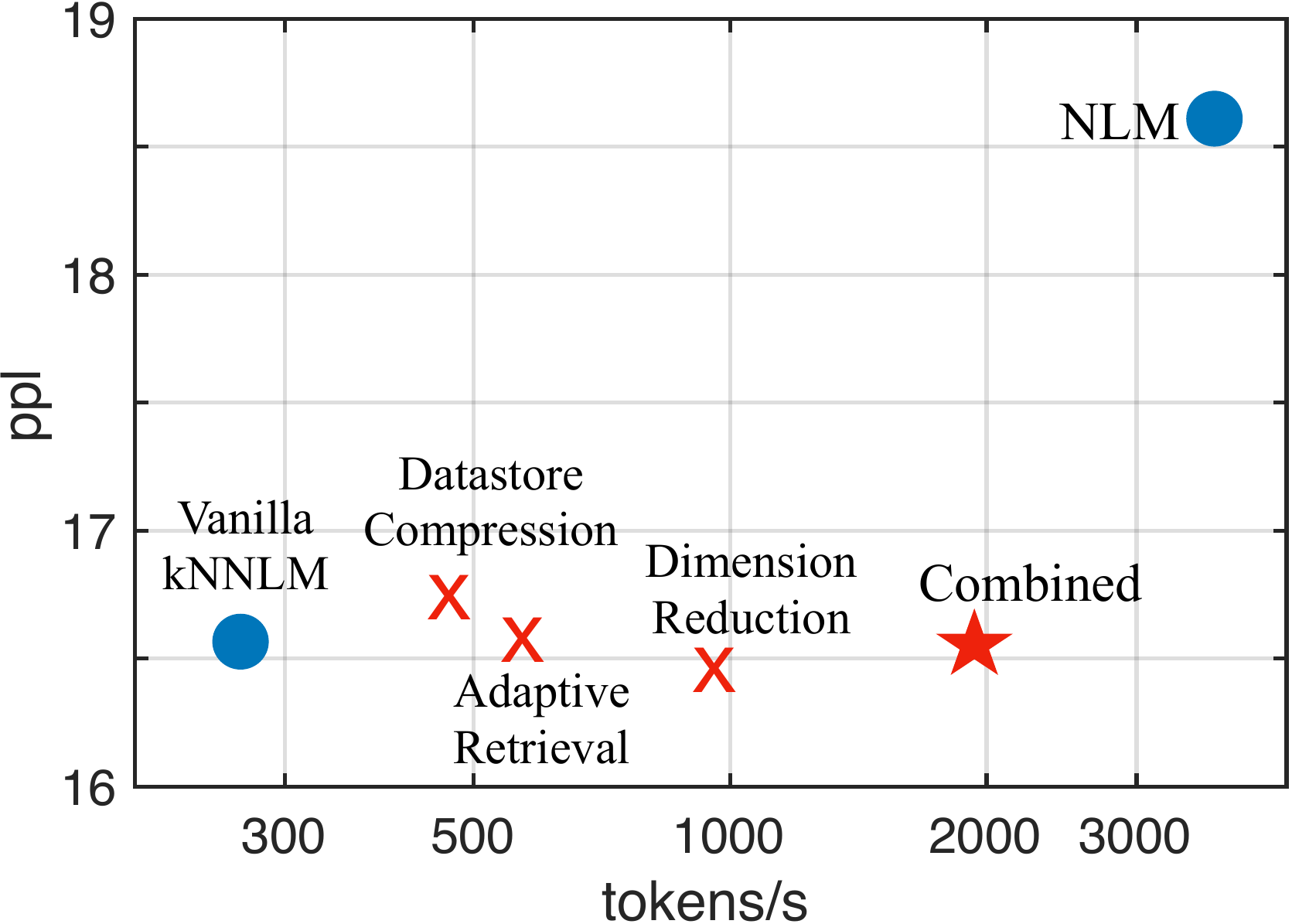}
\caption{\small \label{fig:intro_scatter} Perplexity (ppl) and evaluation speed for different models. Circled points represent the neural language model (NLM) and $k$-nearest neighbors language model (\knnlm) baselines respectively, while others are the methods that we propose and explore in this paper. }
% \vspace{-15pt}
\end{figure}
%such as machine translation~\citep{xx}, summarization~\citep{xx}, dialogue~\citep{xx}, etc. Recently, language modeling is shown to be an effective pretraining technique as well which benefits downstream tasks~\citep{xlnet,gpt} and produces unsupervised multi-task learners~\citep{gpt}. 
Modern neural language models (NLMs) based on recurrent~\citep{mikolov2010recurrent,sundermeyer2012lstm} or self-attentional~\citep{vaswani2017attention,al2019character} neural networks are mostly \emph{parametric}, where the predictions are solely dependent on the model parameters given the input data.

 \begin{SCfigure*}
\centering
\caption{\label{fig:model} Illustration of \knnlm. The datastore consists of paired context representations and the corresponding next tokens. The index represents a method that performs approximate $k$-nearest neighbors search over the datastore. Red and bolded text represent three dimensions that we explore in this paper to improve the efficiency. In adaptive retrieval, we use a predictive model to decide whether or not to query the datastore.}
    \includegraphics[width=0.6\textwidth]{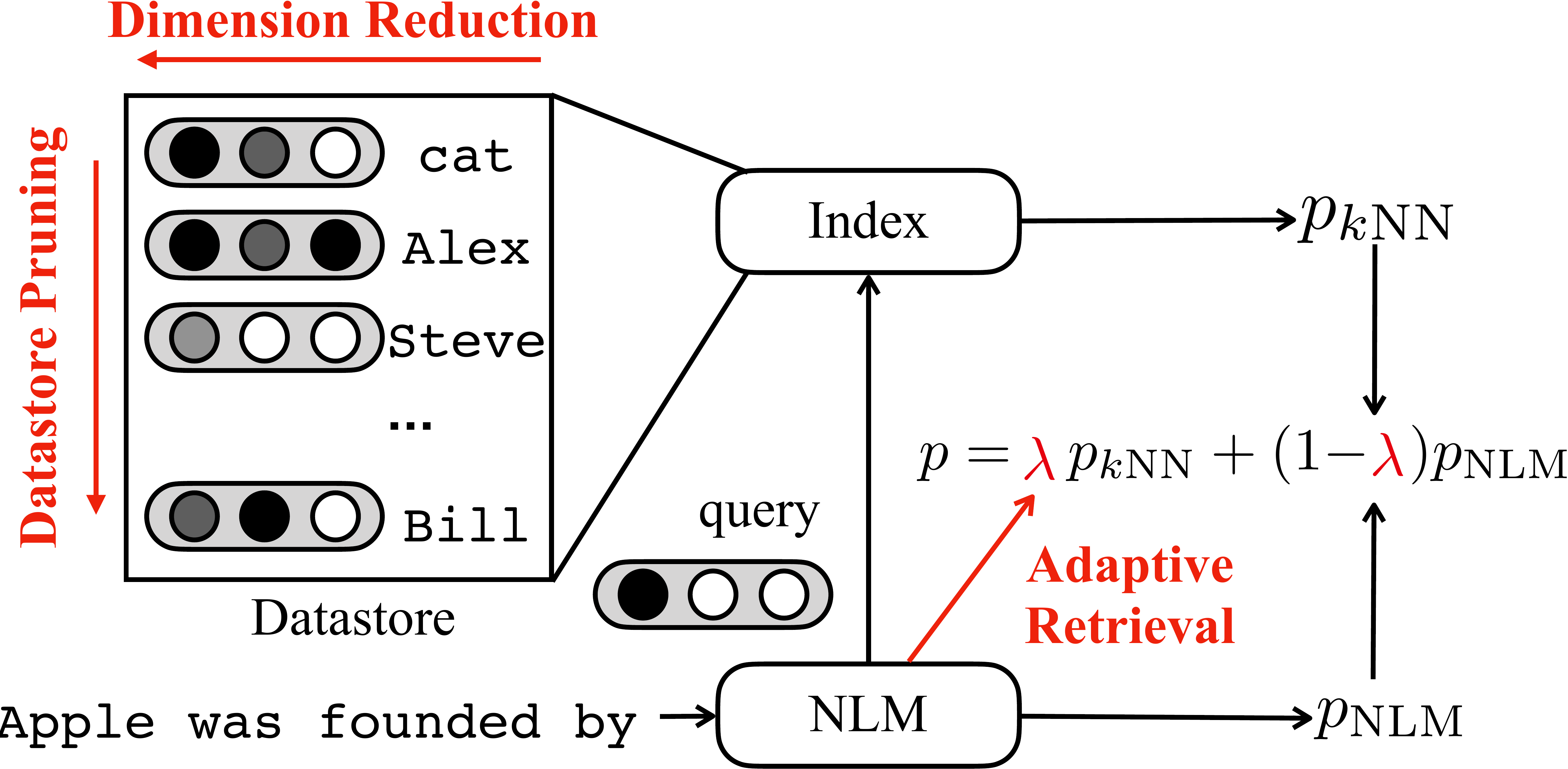}
%\vspace{-15pt}
\end{SCfigure*}

% \small \label{fig:model} Illustration of \knnlm. The datastore consists of paired context representations and the corresponding next tokens. The index represents a method that performs approximate $k$-nearest neighbors search over the datastore. Red and bolded text represent three dimensions that we explore in this paper to improve the efficiency. In adaptive retrieval, we use a predictive model to decide whether or not to query the datastore. \tbk{shrink and then use sidecap package. will save you some space and will look better. could also increase math font size a bit and maybe thicken lines by one point (but only if you shrink it). Could also use courier font for string literals (e.g. next words in datastore and input context)}
In contrast, recent \emph{non-parametric} LMs~\citep{guu2018generating,khandelwal2019generalization,he2020learning} model text distributions by referencing both the parameters of the underlying model \emph{and} examples from an external datastore. 
% \tbk{do we need a caveat that we're specifically describing recent non-parametric neural LMs? `non-parametric' of course has a formal statistical definition, and presumably there are non-parametric LMs that don't use an external datastore, e.g. those based on Bayesian non-parametrics -- just want to be careful here}  
Non-parametric LMs are appealing since they allow for effective language modeling -- particularly for rarer patterns -- through explicit memorization via a datastore, which mitigates the burden on model parameters to learn to encode all information from a large dataset. 
% \gn{I'm pretty skeptical about a number of statements in this sentence. (1) what does ``easily'' mean? maybe ``effectively'' or ``efficiently''? (2) ``generalize to'' might be ``model''? I'm not sure that it's really clear that the models are generalizing better per say, it might just be that parametric models are underfitting? (3) What is an ``unseen pattern'' and is there evidence that non-parametric LMs can model or generalize to them? (4) ``that mitigates the growing burden on the model parameters'' I agree with, but the description could be made more precise (growing with respect to what? why are non-parametric models able to mitigate this burden?)}.
% Among the various recent models that make progress in this direction~\citep{guu2018generating,khandelwal2019generalization,he2020learning}, 
One effective and representative example is the $k$-nearest neighbors LM ($k$NN-LM,~\citet{khandelwal2019generalization}). The $k$NN-LM computes the probability of the next token by interpolating a parametric LM with a distribution calculated from the $k$ nearest context-token pairs in the datastore, as demonstrated in Figure~\ref{fig:model}.
This model is particularly notable for its large improvements in performance -- it outperforms the previous best parametric LMs by a large margin in standard language modeling benchmarks, in domain adaptation settings, and on other conditional generation tasks such as machine translation \citep{khandelwal2020nearest}.
% \gn{Shortened this.}
% Also, $k$NN-LM demonstrates effective performance on domain adaptation and extra training data utilization -- by simply changing the datastore -- without updating model parameters. The same method has been applied to machine translation task and achieved similar success~\citep{khandelwal2020nearest}.
%These successes have been further validated in conditional generation tasks like non-parametric machine translation~\citep{khandelwal2019generalization}.

However, one downside to the $k$NN-LM is that the datastore stores high-dimensional dense vectors for \emph{each token in the training data}; this can easily scale to hundreds of millions or even billions of records.
As a result, the extra retrieval step from such datastores greatly decreases model efficiency at test time.
For example, a 100M-entry datastore can lead to an over 10x slow-down 
% \tbk{..when scoring test sentences? might be a good idea to make this explicit since earlier we use the word `inference' which is a bit ambiguous} 
compared to parametric models (\textsection\ref{sec:speed-baseline}) as shown in Figure~\ref{fig:intro_scatter}. This issue poses a serious hurdle for the practical deployment of non-parametric LMs, despite their effectiveness. 

In this paper, we attempt to address this issue of test-time inefficiency and make non-parametric LMs more applicable in real-world settings.
We take $k$NN-LM as an example, first analyzing the evaluation overhead,
 % \tbk{we are using the word `inference' a lot, and not in a formally correct way -- probably fine, but be a little careful} 
 % cost, and it to the parametric NLMs. Motivated by results of this analysis, 
 and raise three questions that we aim to answer in this paper: (1) Do we really need to perform retrieval on the prediction of every single token? (2) Can we identify and prune redundant records from the datastore? (3) Is it possible to further compress the datastore by reducing the vector dimensionality without losing performance? We propose and explore potential solutions for each question to aid efficiency. Specifically, we (1) show that a lightweight network can be learned to automatically prune unnecessary retrieval operations (adaptive retrieval, \textsection\ref{sec:ar}), (2) explore several different methods for datastore pruning based on clustering, importance-guided filtering, or greedy merging (\textsection\ref{sec:dp}), and (3) empirically demonstrate that simple dimension reduction techniques are able to improve \emph{both} the performance and speed (\textsection\ref{sec:dr}). 
 Figure~\ref{fig:intro_scatter} illustrate the overall performance of these methods. Our experiments on the WikiText-103 language modeling benchmark~\citep{merity2016pointer} and a training-free domain-adaptation setting demonstrate speed improvements of up to 6x with comparable perplexity to the $k$NN-LM.
On a higher level, we expect the empirical results and analysis in the paper to help researchers better understand the speed-performance tradeoff in non-parametric NLMs, and provide a springboard for future research on more efficient non-parametric LMs.

% \tbk{***I noticed when reading the rest of the paper that we don't actually say anything specific about the novel methods we propose until page 5. Perhaps this last paragraph of the intro could provide a brief answer to each of the three questions posed? Stating in very few words each of the three methods we introduce to aid efficiency?***}
% \tbk{intro looks great overall!}

%!TEX root=./emnlp2021.tex

\section{$k$-Nearest Neighbors Language Model}
\label{sec:background}
In this section, we overview $k$NN-LM~\citep{khandelwal2019generalization} and its implementation details. 

\paragraph{Formulation.}
$k$NN-LM~\citep{khandelwal2019generalization} is an LM that estimates token distributions by interpolating a pre-trained autoregressive NLM's distribution with another distribution computed using an external datastore. Specifically, given a sequence of context tokens $c_t = (w_1, \cdots, w_{t-1})$, the $k$NN-LM's next token probability $p(w_t|c_t)$ is calculated through the interpolation of the probability estimated by a standard parametric NLM $p_{\text{NLM}}(w_t|c_t)$ and a probability computed using an external datastore $p_{kNN}(w_t|c_t)$ (detailed later):\footnote{Below, we sometimes ignore the subscript to simplify notation when there is no confusion.}
\begin{equation}
\label{eq:knnlm}
\begin{split}
&p(w_t|c_t) \\
&= \lambda p_{\text{kNN}}(w_t|c_t) + (1-\lambda)p_{\text{NLM}}(w_t|c_t),
\end{split}
\end{equation}
where $\lambda$ is the interpolation hyperparameter.
% $p_{\text{NLM}}$ can be calculated according to any NLM.
Note that the application of $k$NN-LM requires no additional training; the parameters of the NLM remain as-is, and Eq.~\ref{eq:knnlm} is only applied at test time. The workflow of \knnlm is shown in Figure~\ref{fig:model}

\paragraph{Datastore.}
The datastore in $k$NN-LM stores context vectors from the pretrained NLM as keys, and their corresponding next tokens as values. Formally, let $f$ be the key function that maps context sequence $c$ to a fixed-size vector, then the datastore $(\gK, \gV)$ contains all the key-value pairs constructed from the entire training examples $\gD$:
\begin{equation}
(\gK, \gV) = \{(f(c_t), w_t) | (c_t, w_t) \in \gD\}.
\end{equation}
%\tbk{added a parenthesis that was missing from equation above -- make sure I put it in right place. looks right to me}
The size of such a datastore is almost equal to the number of training tokens because the context $c_t$ is (nearly) unique due to the large context window size in modern recurrent~\citep{sundermeyer2012lstm} or self-attentional~\citep{vaswani2017attention} NLMs. This suggests that the datastore can easily scale to hundreds of millions or even billions of records. Also, each $f(c_t)$ is a high-dimensional dense vector, which makes the datastore difficult to fit in memory.
For example, a datastore from a 100M-token training dataset, using 1024-dimension context vectors at 16-bit precision, could require 200GB of memory.%\footnote{Throughout this paper, all the context vectors in the datastore are saved in 16-point float precision to save space.}
\footnote{
%\gn{I moved this to a footnote because it seemed to be a bit of an aside that distracts a little bit from the main point.}
Note that the dataset to contruct the datastore may not necessarily be the training data that trains the parametric NLM in Eq.~\ref{eq:knnlm} -- a separate dataset may be used for the datastore construction which would lead to potential applications such as training-free domain adaptation or a gradient-free way to utilize extra training data~\citep{khandelwal2019generalization}.
}

\paragraph{The Nearest Neighbor Distribution $p_{\text{kNN}}(w_t|c_t)$.}
At inference time, the $k$NN-LM (1) computes the context vector $f(c)$ from the current sequence using the pretrained NLM, (2) uses $f(c)$ as the query to retrieve $k$ nearest neighbors $\gN = \{(q_i, v_i) | i=1,\cdots,k\}$ from the datastore, and (3) aggregates the retrieved tokens to form the distribution $p_{\text{kNN}}(w|c)$ to be used in Eq.~\ref{eq:knnlm} as:
\begin{equation}
\label{eq:pknn}
\begin{split}
& p_{\text{kNN}}(w=y|c) \\
& \propto \sum_{(q_i, v_i)\in\gN}\mathbb{I}_{v_i=y}\exp(-d(q_i, f(c))).
\end{split}
\end{equation}
%\tbk{nice simple equation! this is a better presentation than the original paper perhaps!}
$d(\cdot,\cdot)$ is a distance function between the two vectors, and $L^2$ was shown to be more effective than other alternatives~\citep{khandelwal2019generalization}.
Intuitively, \knnlm finds context sequences in the datastore that are similar to the test context, and then utilizes the next tokens observed after these contexts to help prediction. Such a mechanism allows language modeling through explicit memorization from the datastore, and may be particularly helpful for patterns rarely seen by the pretrained NLM~\citep{khandelwal2019generalization,khandelwal2020nearest}. 

\paragraph{Sources of Inference Overhead.}
%\gn{This section is titled ``Inference Overhead.'' but after reading this section a reader may still not have a good idea of how much inference overhead we're talking about. It seems like it should be mentioned somewhere in this section too. And/or you could change the title to ``Sources of Inference Overhead.''?}
%The test-time workflow of $k$NN-LM is shown in Figure~\ref{fig:model}. 
The extra inference overhead stems from the $k$NN search process in $p_{\text{kNN}}(w_t|c_t)$ computation. We denote the inference time per token as $t = t_{\text{NLM}} + t_{k\text{NN}}$. While $t_{\text{NLM}}$ remains constant with different datasets, $t_{k\text{NN}}$ unfortunately grows as the datastore scales.

In practice, the $k$NN search process is often performed only approximately (ANN, ~\citet{gionis1999similarity,muja2009fast}) to reduce computational cost.
% that finds the nearest neighbors with high probability instead of probability 1 to reduce computational cost. 
%\tbk{worth saying early that we're going to propose additional speedups -- these are just the efficiency tricks the baseline used?} 
\citet{khandelwal2019generalization} implemented ANN search in $k$NN-LM\footnote{\url{https://github.com/urvashik/knnlm}.} using FAISS~\citep{johnson2019billion}, which combines inverted file systems~\citep{sivic2003video} and product vector quantization~\citep{jegou2010product}. This type of index reduces memory usage by only storing quantized vectors and accelerates $k$NN search by pre-clustering the datastore vectors; 
% For instance, this advanced index only requires less than 10GB corresponding to the originally 200GB WikiText-103 datastore. 
interested readers can refer to~\citep{jegou2010product} for more details.
For the purpose of this paper we study \knnlm using this indexing method as a black box, aiming to improve efficiency in an index-agnostic way.
At the same time, we note that building fast and accurate indexing methods remains an active area of research~\citep{andre2019quicker,guo2020accelerating}, and selection or improvement of the index itself (possibly in concert with the methods proposed in this paper) is an interesting avenue for future work.

\paragraph{Distance Recomputation.}
The distances to the nearest neighbors are required to compute $p_{\text{kNN}}(w_t|c_t)$ as shown in Eq.~\ref{eq:pknn}. However, as described above, $k$NN-LM's nearest neighbor search process performs search over quantized vectors, and as a result it can only return \emph{approximate} distances. While it is possible to compute the accurate distances by reading the full-precision vectors from the datastore after retrieval, this presents challenges as well: (1) storing the entire datastore in memory is not scalable for large datastores,
% -- for example, one billion 1024-dimensional vectors would require 2TB memory to fit in, 
(2) reading the vectors from a large datastore on disk on-the-fly is too slow to be practical (< 1 token per second).\footnote{Disk random I/O is another aspect that may be improved by further engineering effort, which is also interesting future work.} Therefore, in this paper we use the approximate distances directly to compute $p_{\text{kNN}}$. This comes at the cost of a minor performance loss, as we will show in~\textsection\ref{sec:speed-baseline}. Similar approximations were adopted to apply $k$NN-LM to machine translation tasks~\citep{khandelwal2020nearest}. 
% In this paper we use approximate distances unless otherwise specified.

%!TEX root=./emnlp2021.tex

\section{The Efficiency of \knnlm}
In this section, we first introduce the datasets and setup that we will use throughout the paper, and then compare the inference speed of \knnlm to parametric NLMs.

\subsection{Datasets}
\label{sec:dataset}
We study \knnlm in two different settings: (1) the standard setting where the datastore is constructed from the same data used to train the NLM, and (2) a domain adaptation setting where the datastore is based on the training data in the test domain, in which case the NLM never sees the examples included in the datastore.
% \gn{The following sentence again seems to not be the main point.}
% Non-parametric NLMs like \knnlm are able to adapt to different domains by simply switching the datastore, without updating neural parameters.
The following two datasets are used for the two settings respectively:

\paragraph{WikiText-103~\citep{merity2016pointer}} is a standard language modeling benchmark from Wikipedia that has 250K word-level vocabulary. It consists of 103M training tokens, and thus leads to a datastore that has 103M records and takes 200G space. Following~\citep{khandelwal2019generalization}, we use the transformer-based~\citep{vaswani2017attention} language model checkpoint released by~\citep{baevski2018adaptive} as the underlying pretrained NLM, which is trained on the WikiText-103 training split.

\paragraph{Law-MT} is an English-German machine translation dataset in the law domain originally released by~\citep{koehn2017six} and resplit by~\citep{aharoni2020unsupervised}. We only use the English text for language modeling. The training set consists of 19M tokens which we use to build the datastore that occupies 55G space. To inspect the domain-adaptation performance, our pretrained NLM is a 12-layer transformer model trained on WMT News Crawl\footnote{\url{http://data.statmt.org/news-crawl/}} released by~\citep{ng2019facebook}. 

\subsection{Setup}
%\gn{I think all of this should come before Section 4, as it's all shared by the experiments in Section 4, right?}
%As our analysis experiments in \textsection\ref{sec:remedy}, 
Throughout the rest of the paper, we adopt the same hyperparameters and index as~\citep{khandelwal2019generalization} for $k$NN-LM.\footnote{We directly base our experiments on the original \knnlm implementation.} 
%Specifically, the index is a mix of an inverted file system and product vector quantization implemented with FAISS~\citep{johnson2019billion}. 
Specifically, the number of nearest neighbors is set to 1024 during evaluation.\footnote{The perplexity continues improving as $k$ grows as shown in~\citep{khandelwal2019generalization} and confirmed by us. Yet $k$ does not have an effect on the evaluation speed in the range [8, 1024] from our observation.} Our pretrained NLMs are the state-of-the-art decoder-only transformers as mentioned above, and the key function $f(c)$ to obtain context vectors is the input to the final layer's feedforward network. The context vectors are 1024-dimensional and 1536-dimensional for WikiText-103 and Law-MT respectively.  Given a dataset, we tune the interpolation weight $\lambda$ on validation set in terms of the vanilla \knnlm performance, and fix it unless otherwise specified. Complete details on the setup can be found in Appendix~\ref{appdix:setup}.
%The retrieval adaptor network in adaptive retrieval is a 5-layer MLP with 128 hidden units. 

Evaluation efficiency is benchmarked on 32 CPU cores (1.5 GHz AMD EPYC 7282) and 1 NVIDIA RTX 3090 GPU which represents a normalized environment -- the index searching uses all the CPU cores while neural network computation is based on the GPU. Running retrieval on 32 CPU cores is also used by the FAISS repo\footnote{\url{https://github.com/facebookresearch/faiss/wiki/Indexing-1G-vectors}} as a standard setting to benchmark large-scale retrieval.

\subsection{Baseline Speed}
\label{sec:speed-baseline}
%\gn{You probably need to discuss hardware, given that these results may be hardware dependent.}
%We follow the same hyperparameters from~\citep{khandelwal2019generalization}, readers can refer to \textsection\ref{xx} for setup details and test environment. 
 We measure the perplexity (ppl) and speed of evaluation in term of tested tokens per second, and Table~\ref{tab:baseline-speed} reports the results on the test set of the two datasets. We also include ``\knnlm (exact)'' for reference, which represents the \knnlm variant that re-computes accurate distances as explained in~\textsection\ref{sec:background}. While very effective with 2 ppl points gains on WikiText-103 and over 90 points gains on Law-MT in a domain-adaptation setting, \knnlm is 10x -- 30x slower to evaluate on these datasets because of the extra retrieval step. When exact distances are computed by reading vectors from the disk on-the-fly, \knnlm (exact) takes over 1 second to evaluate a single token. 
% Next, we perform a case study on WikiText-103 to explore several methods that may potentially improve the efficiency of \knnlm.

\begin{table}[!t]
    \centering
    % \vspace{-0.3cm}
    \caption{Evaluation performance and speed of baseline models. The ppl numbers of \knnlm$^*$ (exact) are from~\citep{khandelwal2019generalization}, which recomputes accurate distances.}
    \label{tab:baseline-speed}
    % \vspace{-0.1cm}
    \resizebox{1 \columnwidth}{!}{
    % \small
    \begin{tabular}{lrrrr}
    \toprule
    %\multirow{2}{*}{\textbf{Model}} & \multicolumn{2}{c|}{\textbf{BLEU}} & \multirow{2}{*}{\textbf{ACC.}} & \multicolumn{2}{c}{\textbf{PPL.}} \\
   \multirow{2}{*}{\textbf{Model}} &  \multicolumn{2}{c}{\textbf{WikiText-103}} & \multicolumn{2}{c}{\textbf{Law-MT}}\\
   &  ppl & tokens/s & ppl & tokens/s\\
    \midrule
\knnlm$^*$ (exact) & 16.12 & <1 & -- & -- \\
\midrule
NLM  &  18.66 & 3847 & 106.25 & 28K \\
\knnlm & 16.65 & 277 & 12.32 & 1052\\
    \bottomrule
    \end{tabular}}
    % \vspace{-10pt}
 \end{table}

\section{The Remedies}
\label{sec:remedy}
In this section we propose and explore several different methods that may potentially improve the efficiency of \knnlm along three axes: (1) adaptive retrieval, (2) datastore pruning, and (3) dimension reduction. We analyze the performance of each method on WikiText-103, trying to conclude the best practices that we will evaluate in \textsection\ref{sec:exp}. 
%Detailed hyperparameters and setup of the analysis in this section can be found in Appendix~\ref{xxx}. 

\subsection{Adaptive Retrieval}
\label{sec:ar}
\begin{table}[!t]
    \centering
    % \vspace{-0.3cm}
    \caption{The features used to train the retrieval adaptor.}
    \label{tab:ar-feature}
    % \vspace{-0.1cm}
    \resizebox{1 \columnwidth}{!}{
    % \small
    \begin{tabular}{ll}
    \toprule
    %\multirow{2}{*}{\textbf{Model}} & \multicolumn{2}{c|}{\textbf{BLEU}} & \multirow{2}{*}{\textbf{ACC.}} & \multicolumn{2}{c}{\textbf{PPL.}} \\
   \textbf{Feature} &  \textbf{Description}\\
   \midrule
   $f(c)$ & \parbox{1.0\columnwidth}{the context embeddings from pretrained NLM} \\
   conf$(c)$ & \parbox{1.0\columnwidth}{the maximal value (confidence) of $p_{\text{NLM}}$} \\ 
   ent$(c)$ & the entropy of the distribution $p_{\text{NLM}}$ \vspace{7pt}\\
   $\log \text{freq}(c[-n:])$ & \parbox{1.0\columnwidth}{log of frequency of the immediate $n$ context tokens computed from the training data. $n=1,2,3,4$ which leads to four scalar features.} \vspace{9pt}\\ 
   $\log \text{fert}(c[-n:])$ & \parbox{1.0\columnwidth}{$\text{fert}(c[-n:])$ is the number of unique word (fertility) that succeeds the immediate $n$ context tokens computed from the training data. $n=1,2,3,4$ which leads to four scalar features.} \\ 
   % $\log $ \\
    \bottomrule
    \end{tabular}}
    % \vspace{-5pt}
 \end{table}
 
 \begin{figure}[!t]
\centering
    \includegraphics[width=1.0\columnwidth]{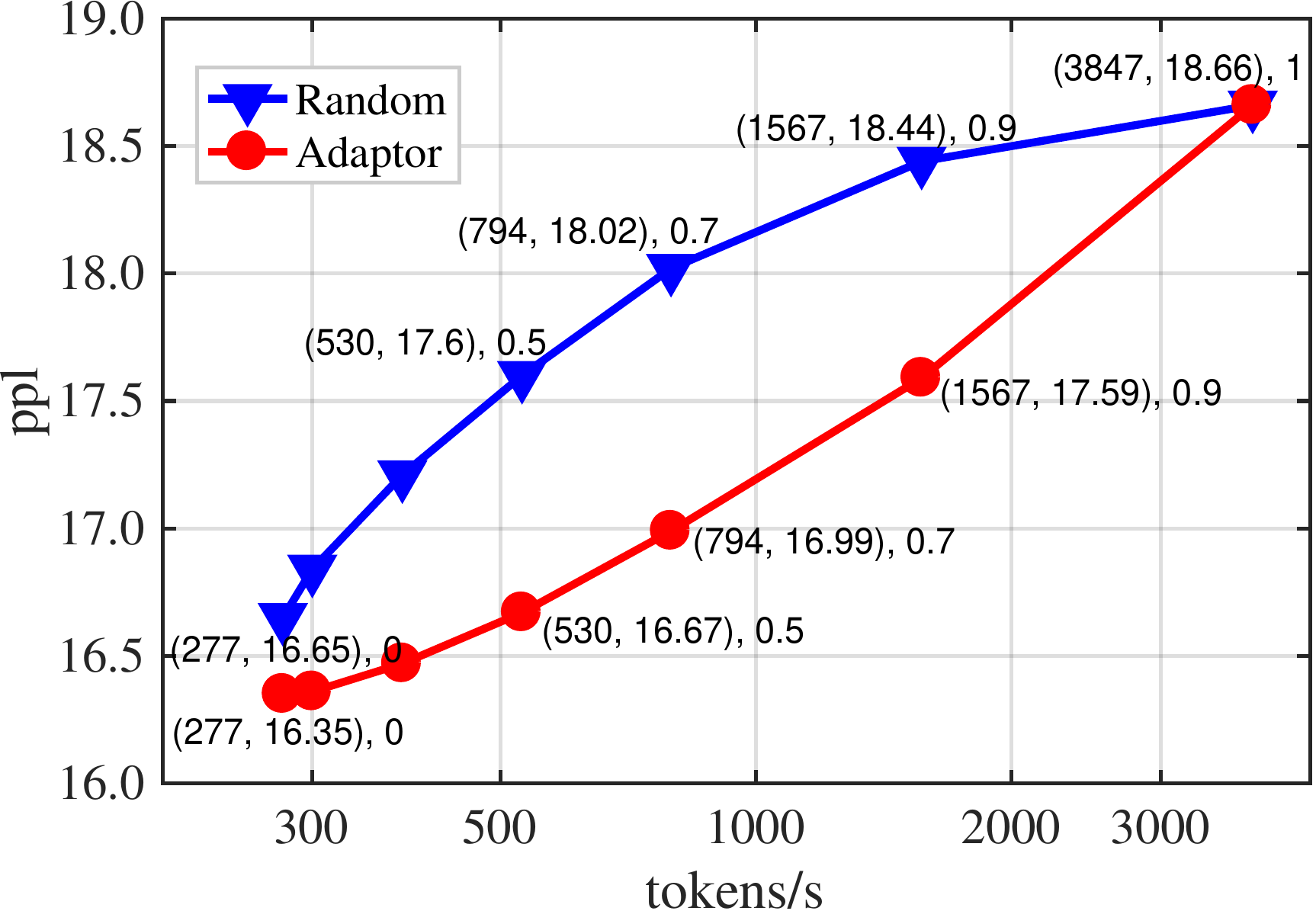}
\caption{\small \label{fig:ar_res} Perplexity and speed results of adaptive retrieval on WikiText-103 test set. We annotate the coordinates of some points and the third number in the annotation is the fraction of retrieval operations that are removed.}
% \vspace{-15pt}
\end{figure}

Just as humans refer to books only when they are uncertain in an open-book quiz, the parametric NLMs may not \emph{always} need help from the external datastore. To inspect this hypothesis, we compare $p_{\text{kNN}}(w|c)$ and $p_{\text{NLM}}(w|c)$ for every token in the WikiText-103 validate set. Interestingly, $p_{\text{kNN}}(w|c) \ge p_{\text{NLM}}(w|c)$ only $39\%$ of the time -- the likelihood of $61\%$ of the tokens becomes worse after interpolation despite the overall improvement. This indicates that if we were able to identify these locations perfectly, $61\%$ of the retrieval operations could be removed completely and we would achieve even better perplexity. Inspired by this observation, we aim to automatically identify and prune unnecessary retrieval operations to speed up inference.
% In the following we describe the methods that we use as well as the results. 

\paragraph{Methods:} 
We propose to train a light neural network, the retrieval adaptor, to identify when we should remove the retrieval operation. Specifically, given the context $c$ as the input, the retrieval adaptor may be trained with either (1) a classification objective to predict whether $p_{\text{kNN}}(w|c) \ge p_{\text{NLM}}(w|c)$, or (2) a likelihood maximization objective to predict the interpolation weight $\lambda(c)$ and maximize the overall likelihood of \knnlm as in Eq.~\ref{eq:knnlm}. In our preliminary results the classification method performs only on par with a random removal baseline, partially due to the discretized noisy supervision. Therefore, we directly maximize the \knnlm log likelihood by modeling $\lambda$ as a function of the context:
\begin{equation}
\gL = \frac{1}{T}\sum_t \big[\log p(w_t|c_t; \lambda_{\theta}(c_t)) - a\cdot\lambda_{\theta}(c_t)\big],
\end{equation}
where only $\theta$ -- the parameters of the retrieval adaptor -- are updated. The second term is an $L^1$ regularizer that encourages learning sparse weights for $p_{k\text{NN}}$, which we find helpful to prune unnecessary retrievals. 
 At inference time, we prune a given fraction of retrievals with the smallest $k$NN weight $\lambda(c)$ by resetting $\lambda(c)$ to zero. The hyperparameters of the retrieval adaptor network including the regularizer coefficient, $a$, are tuned on the validation set in terms of perplexity at 50\% retrieval pruning. Learning the interpolation weights to prune is related to~\citep{johansen-socher-2017-learning} where they learn to skip text for classification tasks. Optimizing the interpolation weights in \knnlm has also been applied at training time to train the NLM jointly~\citep{yogatama2021adaptive}.
 
 \paragraph{Architecture and Input Features:}
%\gn{I feel like at least a little bit of a mention of the model architecture would be useful. In particular, I was wondering what the output layer looked like, was it a sigmoid? This is important because it interacts with L1 regularization. Maybe it could be ``Architecture and Input Features'', maybe move this section before ``Training''?}
The retrieval adaptor is a light MLP network with linear transformation followed by ReLU activation at each layer. The output layer maps the hidden representation to a 2-dimensional vector followed by a \texttt{LogSoftmax} layer to yield $\log(\lambda)$ and $\log(1-\lambda)$ respectively. Complete details  on the retrieval adaptor can be found in Appendix~\ref{appdix:ar}.
We concatenate several neural and count-based features as input to the retrieval adaptor as shown in Table~\ref{tab:ar-feature}. For the scalar features (basically all the features excluding $f(c)$) , we found it helpful to map them to a vector with a small network before concatenation. We note that all the features are trivial to obtain at test time -- the neural features are from intermediate computation of $p_{\text{NLM}}(w|c)$ and count-based features are looked-up values. Ablation analysis on these features can be found in Appendix~\ref{sec:ablation}.

\paragraph{Training:}
During training, only the retrieval adaptor is updated while the pretrained NLM is fixed. Note that it is inappropriate to train the retrieval adaptor on the training dataset, which would lead to biased solutions since $p_{\text{NLM}}$ may have already overfit on the training data and the datastore includes the training example itself. To generalize to the test data, we hold out 10\% of the validation data for validation and use the remaining 90\% to train the retrieval adaptor. The retrieval adaptor is light and converges quickly; it took several minutes to train it on WikiText-103 with a single GPU.

\paragraph{Results:} 
Figure~\ref{fig:ar_res} shows the perplexity and evaluation speed of adaptive retrieval on the test set of WikiText-103, varying the percent of removed retrieval operations. The different threshold values of $\lambda$ used to cut off retrieval is selected based on the synthetic validation set mentioned above. We also add a random retrieval baseline which uniformly selects a certain fraction of retrieval operations to discard. We observe that adaptive retrieval (AR) exhibits a much flatter increase of perplexity than the random baseline when the number of removed retrievals grows.  Notably, AR is able to achieve comparable perplexity to the original \knnlm model (16.67 vs. 16.65) while being nearly 2x faster (530 vs. 277 tokens/s) through removing 50\% of the operations. AR's gain comes from both the smart pruning mask and optimized $\lambda$. We perform an ablation study on this in Appendix~\ref{sec:ablation}.

\subsection{Datastore Pruning}
\label{sec:dp}
The information present in a large training dataset is often redundant, which suggests that a datastore constructed from training tokens may be pruned with no or only minor performance cost. To validate this hypothesis, we propose several different methods to prune the number of entries and reduce the datastore size:

\paragraph{Random Pruning:}
As a simple baseline, a certain fraction of the datastore entries are randomly selected. Random pruning has been shown to work well with a billion-scale datastore in machine translation tasks~\citep{khandelwal2020nearest}. 

\paragraph{$k$-Means Pruning:}
Clustering is a common technique to prune redundant vectors by only keeping the centroids of the clusters. Yet in our task specifically, we note that a general clustering on the context vectors is not directly applicable since the vectors in the same cluster may still correspond to various target tokens, as language use in context is not deterministic. Therefore, we propose to perform target-aware $k$-means clustering -- for a word $w_i$ in the vocabulary, we perform a separate $k$-means clustering for all the context vectors that have $w_i$ as the target token, then we only keep centroids of each cluster as well as saving the cluster size $s$. The \texttt{(centroid vector, cluster size, target token)} triples form a new compressed datastore. Since we approximate multiple vectors in the same cluster with the centroid and only save the centroid vector once in the new datastore, the computation of the $k$NN distribution $p_{k\text{NN}}$ needs to be rectified as:
\begin{equation}
\label{eq:knn-kmeans}
\begin{split}
&p_{\text{kNN}}(w=y|c) \\
&\propto \sum_{(q_i, v_i)\in\gN}\mathbb{I}_{v_i=y} \ s_i\cdot\exp(-d(q_i, f(c))),
\end{split}
\end{equation}
the cluster size $s_i$ acts like weights for each datastore entry. Eq.~\ref{eq:knn-kmeans} recovers Eq.~\ref{eq:knnlm} when every cluster is of size 1.% 
\footnote{In addition, the centroid formulation is roughly equivalent to saving vectors within the same cluster as the centroids multiple times without pruning in the original formulation.}
%However, this is not strictly equivalent since the cluster centroid would be counted as $s$ neighbors in the original formulation but only one neighbor in Eq.~\ref{eq:knn-kmeans} while the total number of retrieved neighbors is fixed the same.} 
In practice, we perform 5000 separate $k$-means clustering passes only for the most frequent 5000 words due to high computational cost, which accounts for 84\% of all the training tokens. For other vectors we treat each of them as a separate clusters with size 1. The number of clusters in $k$-means are set to 1/20 of the number of vectors to be clustered, which produces a 5x smaller datastore overall. We did not intensively tune the $k$-means hyperparameters due to the computational burden. We note that the clustering here is different from the pre-clustering in the ANN index with inverted file systems mentioned in \textsection\ref{sec:background}-- the index's pre-clustering does not actually reduce size and is just for lookup.
%\tbk{Worth adding a quick sentence about how this kind of clustering is different from the pre-clustering that already happens in the fast index (ANN)? i.e. ANN's clustering doesn't actually reduce size -- it's just for lookup. Maybe unnecessary to say, or maybe could just be footnote}

\paragraph{Greedy Merging:}
%\gn{Given that this is so close to $k$-Means it seems like this should be discussed before rank-based pruning?} \tbk{agreed}
\begin{algorithm}
\small
\centering
\caption{Greedy Merging}
\label{alg:gm}
\begin{algorithmic}[1]
% \State $K$
\State $(\gK, \gV)=\{(q_i, v_i)\}_{i=1}^N \leftarrow$ the old datastore
% \State $(\gK^{\prime}, \gV^{\prime}, \gS^{\prime}) \leftarrow$ Initialize a new datastore with added weight field $\gS$
\State $s \leftarrow \bm{1}$ \Comment{weight vector with size $N$}
% \State $aggressive \leftarrow$ TRUE
\For{$(q_i, v_i) \in (\gK, \gV)$}
    \State retrieve $K$ neighbors of $(q_i, v_i)$ as $\{q_{t_k}, v_{t_k}\}^K_{k=1}$ % \in (\gK, \gV)$  
    % \State $T=\{t_k\}_{k=1}^K \leftarrow$ positions of the nearest neightbors in $(\gK, \gV)$ 
    \For{$k = 1,2,\cdots, K$}
        \If{$s_{t_k}=1$ \& $v_i = v_{t_k}$ \& $t_k \neq i$}  \Comment{merge condition}
            \State $s_i \leftarrow s_i + 1$ \Comment{merge $(q_{t_k}, v_{t_k})$ into $(q_i, v_i)$}
            \State $s_{t_k} \leftarrow s_{t_k} -1$ \Comment{Remove record $t_k$}
        \EndIf
    \EndFor
\EndFor
\State Save datastore $\{(q_i, v_i, s_i) | s_i >0, i=1,\cdots,N\}$ 
\end{algorithmic}
\end{algorithm}
Generally we aim to merge records that share the same target token while being close to each other in vector space. Token-aware clustering is an attempt to achieve this goal, but forcing all points to participate in clustering -- and the resulting large clusters -- causes some points within the same cluster to be distant in some clusters with high variance.
Thus approximating all the vectors with the cluster centroids may lead to large errors.
To address this issue, we propose a simple approach, greedy merging (GM), which inspects every record in the datastore and greedily merges their nearest neighbors if a merging condition is satisfied. The detailed algorithm is shown in Algorithm~\ref{alg:gm}. Intuitively, GM is density-based to group points with nearest neighbors, but the merging operation only happens \emph{locally} between a point and its nearest neighbors -- it never propagates to merge the nearest neighbors of nearest neighbors unlike typical density-based clustering methods~\citep{esterdensity} which may amplify errors. 
%\gn{I don't really understand what you're referring to here... $k$-means clustering doesn't work by merging nearest neighbors of nearest neighbors either. It just works by finding the closest centroid, right?}\tbk{Graham is right -- though I guess if we were also learning the variance in k-means it could sort of do what you say, Junxian}. 
Similar to $k$-means pruning, we also compute the weights $s_i$ of each entry in the compressed datastore to correct $p_{k\text{NN}}$ computation using Eq.~\ref{eq:knn-kmeans}. Without a global clustering mechanism, this approach ensures that the merging vectors are close enough by inspecting only a small number of nearest neighbors. In the following analysis we vary the number of nearest neighbors $K$ within range [2,100] to achieve different compression rates. 

\paragraph{Rank-based Pruning:}
It is well known that embedding spaces contain ``hubs'' which are nearest neighbors of many other embeddings~\citep{tomasev2013role}, and other points that are not nearest neighbors of any other points.
We hypothesize that these entries which are rarely nearest neighbors may be removed without significant impact on the performance. To verify this assumption, we iterate every $(c_i, w_i)$ pair in the training data as queries to search their $k$ nearest neighbors from the datastore ($k$ is set to a large number as 1024 here). In this process we compute an ``importance score'' for every entry in the datastore as $g = \sum_i 1/\text{rank}_i$, where $\text{rank}_i$ is the rank of this entry among the nearest neighbors of the query $f(c_i)$. rank$=+\infty$ if it is not in the retrieval results. Intutively, the ``importance score'' up-weights the datastore records that appear more often with lower ranks in the retrieval results. Then we sort all the datastore records in terms of $g$ and remove the ones with small scores, varying the compression rate. This method shares spirit with the technique in~\citep{min2021neurips} which filters out the articles that are never retrieved in memory-constrained open-domain question answering tasks.

\begin{figure}[!t]
\centering
    \includegraphics[width=1.0\columnwidth]{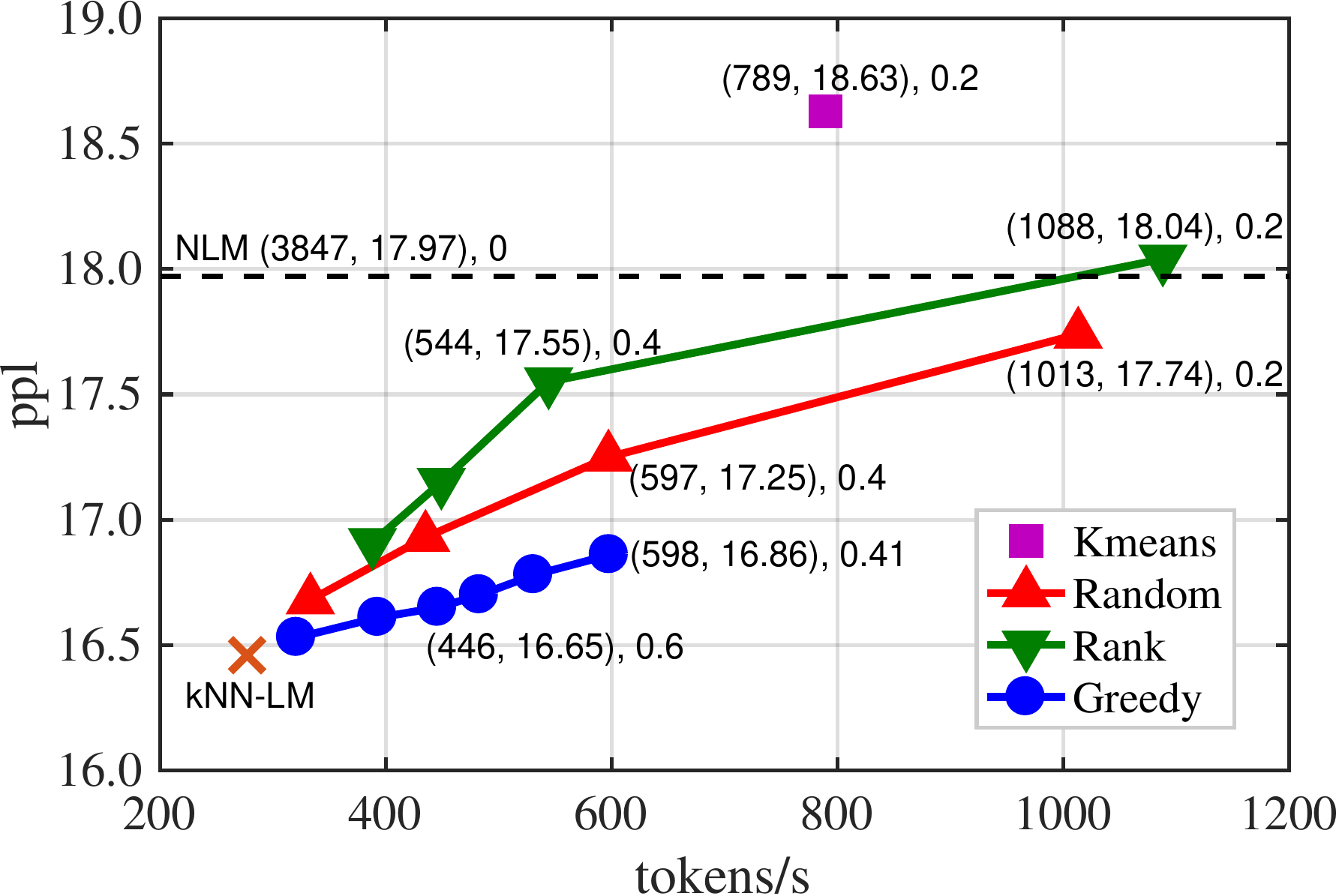}
\caption{\small \label{fig:dp_res} Perplexity and speed results of datastore pruning methods on WikiText-103 validation set. We annotate the coordinates of some points and the third number in the annotation is the compression rate (fraction of records remained).}
%\vspace{-15pt}
\end{figure}

% Greedy merging is theoretically sound as well by upper-bounding the distance bias at inference time -- suppose the distance between the test-time query and the original datastore vector $q_1$ is $d_1$, and $q_1$ is merged to $q_2$ after greedy merging which yeilds a new distance $d_2$, then the distance bias $||$
\paragraph{Results:}
Figure~\ref{fig:dp_res} demonstrates the perplexity v.s. speed results on Wikitext-103 validation set of different datastore pruning methods described above. Only one solution point is reported for $k$-means since we do not vary the hyperparameters of $k$-means for different compression rate, given that its computational cost is much higher than other methods. Using 20\% of the original datstore, $k$-means even underperforms the vanilla NLM baseline, suggesting that the cluster centroids approximation may lead to large distance errors which reduce the accuracy of the $k$NN distribution. Surprisingly, the simple random pruning method outperforms more complicated ones such as $k$-means and rank-based pruning. The best approach is greedy merging, which demonstrates a relatively flat curve compared with others.
%\gn{This is indeed surprising. It'd be nice if you could explain it a bit more.}.
%In addition to the speed-up afforded by adaptive retrieval, the datastore pruning techniques in this section are able to be used to prune redundant information in a text dataset, which may have implications for other applications.
% \gn{such as?}.

\subsection{Dimension Reduction}
\label{sec:dr}
\begin{figure}[!t]
\centering
    \includegraphics[width=1.0\columnwidth]{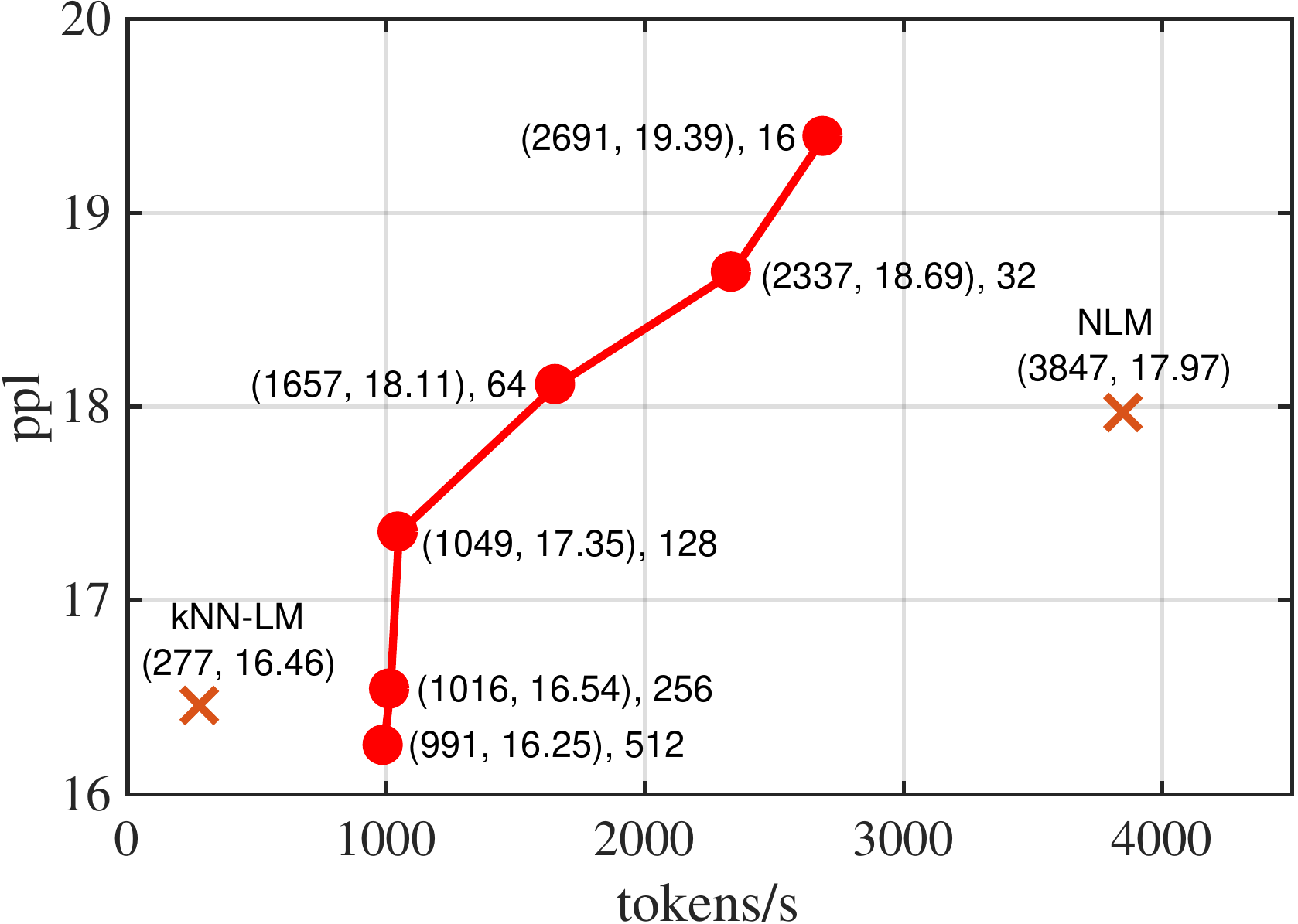}
\caption{\label{fig:pca_res} PCA dimension reduction results on WikiText-103 validation set. We annotate the coordinates of the points and the third number in the annotation is the PCA dimensions.}
%\vspace{-15pt}
\end{figure}
The context vectors $f(c)$ from large NLMs are often high-dimensional.
For example, the pretrained NLMs that we use produce vectors of 1024 and 1536 dimensions in WikiText-103 and Law-MT respectively, which incurs significant datastore space and distance computation cost. To mitigate this issue, we empirically explore the effect of dimension reduction in \knnlm. Specifically, we use principal component analysis (PCA), an efficient and scalable dimension reduction algorithm, to reduce the dimensions and generate a new compressed datastore. We vary the new PCA dimensions as the hyperparameter and report the results.
\paragraph{Results: }
As shown in Figure~\ref{fig:pca_res}, the evaluation becomes faster as expected with smaller dimensions, yet a too aggressive compression (dimension < 256) incurs large perplexity cost and even loses advantages over NLM when the dimension is smaller than 128. However, at 256 and 512 dimensions PCA is able to achieve comparable or even better  performance than the original 1024-dim vectors, while attaining 3x-4x speed-up.\footnote{The tool we use for PCA, the FAISS PCA implementation, applies random rotation to the PCA output vectors by default to re-balance variances of components of a vector~\citep{gong2012iterative}, which may provide additional benefits over vanilla PCA on product vector quantization inside the index.}

%!TEX root=./emnlp2021.tex
\section{Putting it All Together}
\label{sec:exp}
Based on the analysis results in \textsection\ref{sec:remedy}, in this section we combine best practices in adaptive retrieval, datastore pruning, and dimension reduction to assess the performance.
%We also present additional ablation analysis.
We select the retrieval pruning rate $r$, datastore pruning rate $n$, and the reduced dimensions $d$ on the validation set,\footnote{Adaptive retrieval uses part of the validation data to training the retrieval adaptor network, thus we select $r$ separately on its own held-out validation and then combine it to others.} so that they achieve the largest speed-up at the cost of $<=0.1$ perplexity compared to vanilla \knnlm. We report the results on the test set. 
\paragraph{Results:}
\label{sec:exp-results}
\begin{table}[!t]
	\centering
	% \hspace{-0.25cm}
	\caption{Perplexity and speed results on the test set of WikiText-103 and Law-MT. AR, GM, DR denote adaptive retrieval, datastore pruning, and dimension reduction respectively, ``+All'' denotes the combination of all the three technique.}
	\resizebox{1.0 \columnwidth}{!}{
	\begin{tabular}{lrrr}
	\toprule
	\textbf{Methods} & ppl & tokens/s & speedup \\
	\hline
    \multicolumn{4}{c}{WikiText-103} \vspace{0.7mm}\\
    %\hline
        NLM & 18.66 & 3847 & 13.9x\\
        \hdashline
    	\knnlm & 16.65 & 277 & 1x \\
    	 \ +AR ($r=0.5$) & 16.67 & 530 & 1.9x \\
    	 \ +GM ($n=0.6$) & 16.86 & 446 & 1.6x \\
    	 \ +DR ($d=512$) & 16.40 & 991 & 3.6x \\
    	 \ +All & 16.67 & 1835 & 6.6x \\
	\midrule
    \multicolumn{4}{c}{Law-MT} \vspace{0.7mm}\\
    %\hline
        NLM & 106.56 & 27.8K & 264.3x\\
        NLM (fine-tuned) & 8.61 & 27.8K & 264.3x\\
        \hdashline
    	\knnlm & 12.64 & 1052 & 1x \\
    	 \ +AR ($r=0.1$) & 12.74 & 1290 & 1.2x \\
    	 \ +GM ($n=0.6$) & 13.33 & 1451 & 1.4x \\
    	 \ +DR ($d=512$) & 11.59 & 3420 & 3.3x \\
    	 \ +All & 12.29 & 5708 & 5.4x \\
	\bottomrule 
	\end{tabular}}
	\label{tab:main-res}
 	\vspace{-10pt}
\end{table}

Table~\ref{tab:main-res} shows the results on the test set of WikiText-103 and Law-MT, where we assess the combination of all three different strategies. Separate performance for each strategy is also included for reference points. On WikiText-103, adaptive retrieval is able to remove 50\% of the retrieval and achieve nearly 2x speed-up, greedy merging prunes 40\% of the datastore at the cost of 0.2 perplexity points. The dimension reduction method PCA leads to a minor improvement of perplexity over \knnlm while being 3.6x faster. Combination of all the three techniques yields comparable perplexity to vanilla \knnlm (16.67 v.s. 16.65) and a 6.6x speed-up (1835 v.s. 277). 

Different from WikiText-103 where the datastore is contructed from the data that trains the pretrained NLM, in the Law-MT domain adaptation setting the datastore represents the domain-specific knowledge that the pretrained NLM never sees during training and thus is critical to produce good perplexity. This may be inferred from by the large ppl gains that the datastore offers (94 points). From another perspective though, the big improvement from the datastore retrieval leads to difficulties removing retrieval operations adaptively\footnote{This can be reflected from the oracle comparison: $p_{\text{kNN}}(w|c) \ge p_{\text{NLM}}(w|c)$ 76\% of the time compared to 39\% in WikiText-103.} --  our learned retrieval adaptor is able to remove only 10\% of the retrieval operations costing 0.1 ppl points. Greedy merging is able to prune 40\% of the datastore losing 0.7 ppl points. We suspect that the Law-MT datastore is more vulnerable to pruning than the WikiText-103 one because of its smaller size (19M v.s. 103M) and corresponding lack of redundancy. 
% A billion-level or larger datastore in a specific domain may be robust even to random pruning, as shown in machine translation tasks~\citep{khandelwal2020nearest}. 
Interestingly, the PCA dimension reduction yields 1 point ppl gain over the vanilla \knnlm while achieving 3.3x speed-up, consistent with WikiText-103. This implies that a PCA transformation may be able to produce a new vector space that is more appropriate for defining $p_{k\text{NN}}$ with $L^2$ distances, we leave the underlying reasons for future work to discuss. Finally, a combination of the three allows \knnlm to be evaluated 5.4x faster and even obtain superior perplexity.

\section{Implications and Future Work}
% \gn{Changed the title, as ``conclusions'' are often boring rehashes of the content of the paper.}\tbk{agree with Graham on what the content of this section should say (i.e. not just a boring rehash), but i'd lean towards conventional title of `conclusion' -- though fine with either, up to you Junxian}
In this paper, we explore several different ways to improve efficiencies of the $k$-nearest neighbors language model, achieving up to 6x speed-up while attaining comparable performance. As for future work, it is interesting to explore features from the datastore side to better know when to retrieve, and the gap between retrieval-based NLMs and parametric NLMs may be further reduced by combining more optimized indexing methods and the approaches in this paper.
\section*{Acknowledgements}
We thank the anonymous reviewers for their comments, Emma Strubell, André Martins, Pedro Martins, and Uri Alon for helpful advice and discussions, and Wanzhen He for help with figure plotting. This material is based upon work supported by the National Science Foundation under Grant 1815287.

\bibliography{anthology,custom}
\bibliographystyle{acl_natbib}

\appendix

\clearpage
%!TEX root=./emnlp2021.tex
\newpage
\section{Experimental Setup Details}
\label{appdix:setup}
\subsection{General Setup Details}
The interpolation hyperparameter $\lambda$ is tuned in the range [0,1, 0.9] with interval 0.05 on the validation split of each dataset separately. As a result, $\lambda=0.25$ in WikiText-103 and $\lambda=0.9$ in Law-MT.

\subsection{Adaptive Retreival}
\label{appdix:ar}
We use the same adaptive retrieval configuration hyperparameters for different datasets, which are validated on the WikiText-103 dev set: the retrieval adaptor is a MLP network with 4 hidden layers, 1 input layer and 1 output layer. Each layer is a linear transformation followed by the ReLU non-linear activation, and a dropout layer with 0.2 dropout probability, except for the output layer where the hidden unites are transformed to 2 dimensions followed by a log softmax to produce $\log\lambda$ and $\log(1-\lambda)$. The number of hidden units in each layer is 128. Before passing the input features to MLP, we transform each of the scalar features (all the features except for $f(c)$) into an $m$-dim vector, where $m=\text{dim}(f(c)) / n$ and n is the number of scalar feature types. This is to balance the context vector feature and other features. The scalar-feature transformation is performed with an one-layer \texttt{Linear(in, out)-ReLU-Linear(out, out)} network. We also tried using LSTM~\citep{hochreiter1997long} network to capture the temporal relations yet found it leads to very unstable training and fails to converge, though we note that MLP is faster at test time and the $f(c)$ feature already captures the temporal correlations between tokens. The coefficient of the $L^1$ regularizer $a$ is tuned on WikiText-103 validation set among $\{0.01, 0.05, 0.1, 0.2, 0.5, 1\}$ and fixed as 0.05 for both WikiText-103 and Law-MT. The model is trained using the Adam optimizer~\citep{kingma2014adam} with learning rate 0.0005. The checkpoint with the best validation perplexity at 50\% pruning is saved.

\section{Ablation Analysis}
\label{sec:ablation}
\paragraph{Input features to retrieval adaptor:}
We analyze the effect of different input features to the retrieval adaptor by removing a subset of features. We report the perplexities at 50\% retrieval pruning, because using different features only has a marginal effect on the evaluation speed. Results on the WikiText-103 test set are shown in Table~\ref{tab:ablation-feature}. All features together produce the best results, while the perplexity is relatively robust to removal of a single feature. In our experiments (\textsection\ref{sec:remedy} and \textsection\ref{sec:exp-results}) we drop off the $\log\text{freq}$ feature and use the others to save memory while achieving comparable perplexities to using all features.
\begin{table}[!t]
    \centering
    % \hspace{-0.25cm}
    \caption{Results of \knnlm + adaptive retrieval using different input features. The perplexities are based on removing 50\% of the retrieval operations after training retrieval adaptor.}
    \resizebox{0.8 \columnwidth}{!}{
    \begin{tabular}{lr}
    \toprule
    \textbf{Features} & \textbf{ppl}  \\
    \midrule
    %\hline
        $f(c)$ + conf + ent + $\log\text{freq}$ + $\log\text{fert}$  & 16.63 \\
        \ \ -$\log\text{freq}$ & 16.67 \\
        \ \ -$\log\text{fert}$ & 16.72 \\
        \ \ -ent, conf & 16.77 \\
        \ \ -$f(c)$ & 16.71 \\
        \ \ -conf, ent, $\log\text{freq}$, $\log\text{fert}$ & 17.03 \\
    \bottomrule 
    \end{tabular}}
    \label{tab:ablation-feature}
    % \vspace{-7mm}
\end{table}

 \begin{figure}[!t]
\centering
    \includegraphics[width=1.0\columnwidth]{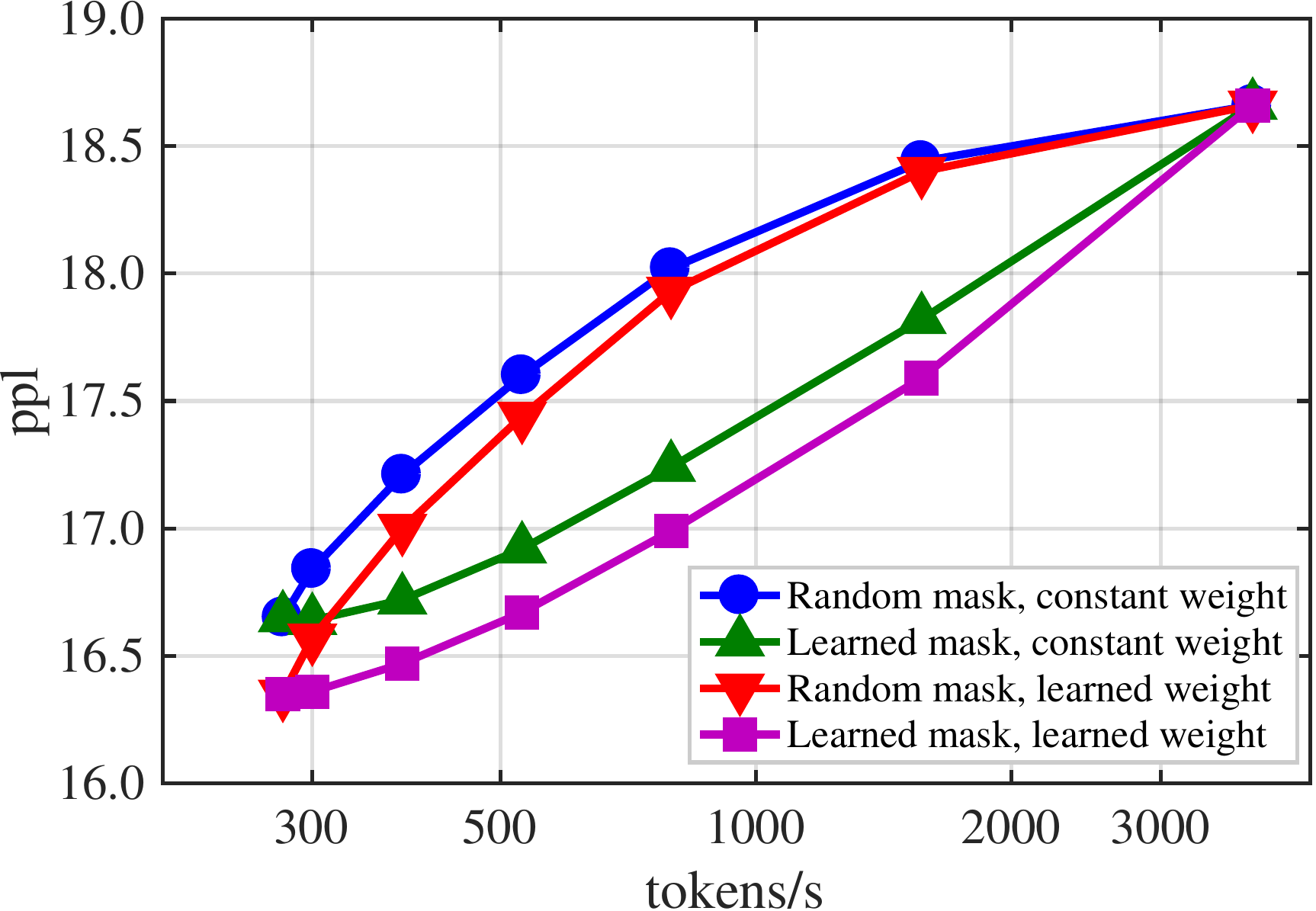}
\caption{\small \label{fig:ar-ablation} Perplexity and speed results of adaptive retrieval on WikiText-103 test set. This figure includes different variants of adaptive retrieval for ablation analysis.}
%\vspace{-15pt}
\end{figure}

\paragraph{Effect of learnable interpolation weights:}
In the adaptive retrieval analysis (\textsection\ref{sec:ar}), we observed gains of a learned retrieval adaptor over a random baseline at different fractions of retrieval prunning. However, the advantages may come from two sources: (1) the automatically identified prunning masks agains the random masks, and (2) the learned interpolation weights on the remaining retrievals against the constant weights that random baseline uses. To separate the two effects, we perform an ablation study to analyze the results of (1) random mask, constant weight (``Random'' in \textsection\ref{sec:ar}), (2) random mask, learned weight -- the weights are from the trained retrieval adaptor, (3) learned mask, constant weight, and (4) learned mask, learned weight (``Adaptor'' in \textsection\ref{sec:ar}). The results are shown in Figure~\ref{fig:ar-ablation}, ``learned mask, learned weight'' performs the best. While minor gains are from the automatically learned weights (``Random mask, learned weights''), most of the superiority can be attained with the smart pruning strategy even with constant weights (``Learned mask, constant weights'').

\end{document}